\documentclass[final, 5p,times,twocolumn, preprint]{elsarticle}
\usepackage{amsmath, amssymb, amsfonts}
\usepackage{algorithm}
\usepackage{algpseudocode}
\usepackage{enumerate}
\usepackage{tabularx}
\usepackage{textcomp}
\usepackage{amsthm}
\usepackage{booktabs}
\usepackage{multirow}
\usepackage{ulem}
\usepackage{tabularx}
\usepackage{threeparttable}
\usepackage{caption}
\usepackage{makecell}
\usepackage[switch]{lineno}
\usepackage{color}
\usepackage[table]{xcolor}
\usepackage{url}
\usepackage{bbding}
\newcolumntype{L}[1]{>{\raggedright\arraybackslash}p{#1}}
\newcolumntype{C}[1]{>{\centering\arraybackslash}p{#1}}
\newcolumntype{R}[1]{>{\raggedleft\arraybackslash}p{#1}}

\newcommand{\clr}[1]{{\color{black}{#1}}}
\newcommand{\clred}[1]{{\color{black}{#1}}}
\newcommand{\clredd}[1]{{\color{black}{#1}}}

\newcommand{\op}[1]{\operatorname{#1}}
\newcommand{\egi}{\textit{e.g.}}
\journal{Biomimetic Intelligence and Robotics}

\begin{document}
\begin{frontmatter}

\linenumbers

\title{HAPNet: Toward Superior RGB-Thermal Scene Parsing via \\ Hybrid, Asymmetric, and Progressive Heterogeneous Feature Fusion}

\author[label1]{Jiahang Li}\ead{lijiahang617@tongji.edu.cn}
\author[label2]{Peng Yun}\ead{pyun@connect.ust.hk}
\author[label3]{Yang Xu}\ead{xuyang0098@smbu.edu.cn}
\author[label3]{Ye Zhang}\ead{ye.zhang@smbu.edu.cn}
\author[label4,label6,label7,label8]{Mingjian Sun}\ead{sunmingjian@hit.edu.cn}
\author[label1]{Qijun Chen}\ead{qjchen@tongji.edu.cn}
\author[label5]{Ilin Alexander}\ead{iline@cs.msu.su}
\author[label1]{Rui Fan\corref{cor}}\ead{rui.fan@ieee.org}

\affiliation[label1]{organization={College of Electronic and Information Engineering, Tongji University}, 
            city={Shanghai}, postcode={201804}, country={China}}

\affiliation[label2]{organization={Department of Computer Science and Engineering, Hong Kong University of Science and Technology}, 
            city={Hong Kong}, country={China}}

\affiliation[label3]{organization={MSU-BIT-SMBU Joint Research Center of Applied Mathematics, Shenzhen MSU-BIT University}, 
            city={Shenzhen}, postcode={518172}, country={China}}

\affiliation[label8]{organization={Qingdao Innovation and Development Base, Harbin Institute of Technology (Weihai)}, 
            city={Qingdao}, postcode={266109}, country={China}}

\affiliation[label4]{organization={Department of Control Science and Engineering, Harbin Institute of Technology}, 
            city={Harbin}, postcode={150001}, country={China}}

\affiliation[label6]{organization={Harbin Institute of Technology at Weihai}, 
            city={Weihai}, postcode={264200}, country={China}}

\affiliation[label7]{organization={Suzhou Research Institute, Harbin Institute of Technology}, 
            city={Suzhou}, postcode={215000}, country={China}}

\affiliation[label5]{organization={Faculty of Computational Mathematics and Cybernetics, Lomonosov Moscow State University}, 
            city={Moscow}, postcode={119991}, country={Russia}}
            
\cortext[cor]{R. Fan is the corresponding author.}

\begin{abstract}
Data-fusion networks have shown significant promise for RGB-thermal scene parsing. However, the majority of existing studies have relied on symmetric duplex encoders for heterogeneous feature extraction and fusion, paying inadequate attention to the inherent differences between RGB and thermal modalities. \clr{Recent progress in vision foundation models (VFMs), which leverage self-supervised learning on large-scale unlabeled datasets, has exhibited superior capabilities in extracting informative, general-purpose features compared to supervised encoders.} However, their potential has yet to be fully leveraged in the domain. In this study, we take one step toward this new research area by exploring a feasible strategy to fully exploit VFM features for RGB-thermal scene parsing. \clr{Specifically, we delve deeper into the unique characteristics of RGB and thermal modalities, thereby designing a hybrid, asymmetric encoder that incorporates both a VFM and a cross-modal spatial prior descriptor (CSPD), enabling enhanced extraction of complementary heterogeneous features. The extracted features undergo dual-path feature fusion through our proposed progressive heterogeneous feature integrators. Moreover, we introduce an auxiliary task to further enrich the local semantics of fused features, thereby improving the overall performance of RGB-thermal scene parsing.} \clr{Our proposed HAPNet, incorporating all these components, delivers superior performance under challenging illumination conditions. }Extensive experiments demonstrate that HAPNet outperforms all other state-of-the-art methods, with improvements of 0.1\%, 1.0\%, and 2.4\% in mIoU on three public RGB-thermal scene parsing datasets: MFNet, PST900, and KP Day-Night, respectively. Additionally, our method exhibits exceptional generalizability for RGB-HHA scene parsing. We believe this new paradigm has opened up new opportunities for future developments in data-fusion scene parsing approaches.
\end{abstract}

\begin{keyword}
data-fusion, thermal, scene parsing, heterogeneous feature, vision foundation model.
\end{keyword}

\end{frontmatter}

\section{Introduction}
\label{sect.intro}

\subsection{Background}
Unity in diversity strengthens perception. RGB-thermal (often abbreviated as RGB-T) scene parsing has emerged as a crucial feature in autonomous vehicles and mobile robots \cite{shivakumar2020pst900}. RGB images capture visible light, while thermal images capture heat signatures. The fusion of these two modalities of data, especially during the feature encoding stage, has been proven to dramatically enhance the reliability and robustness of scene parsing \cite{liang2023explicit, sun2019rtfnet, ha2017mfnet}. Nevertheless, current data-fusion approaches \cite{zhang2021abmdrnet, guo2025lix, shin2023complementary, zhang2023cmx, zhou2021gmnet, lv2024context, s3mnet, zhou2023cacfnet, depthmatch} generally employ duplex \cite{he2016deep, liu2021swin, ticoss, liu2022convnet, lee2025sg} encoders (two identical pre-trained hierarchical backbone networks with independent trainable parameters) to indiscriminately extract heterogeneous features from the RGB-T data, limiting their ability to fully exploit the advantages of both data modalities \cite{feng2024sne}. Therefore, this study aims to explore a more ingenious and effective strategy for heterogeneous feature extraction and fusion, with a specific focus on leveraging VFMs, as exemplified by BEiT series \cite{bao2021beit, peng2022beitv2} and DINOv2 \cite{oquab2023dinov2}, which have garnered considerable attention in recent computer vision research endeavors.

\subsection{Existing Challenges and Motivation}
The backbone networks in duplex encoders are commonly pre-trained on the ImageNet database \cite{krizhevsky2012imagenet} (containing millions of annotated natural images) through fully supervised learning. Although the ImageNet database has enabled these networks to closely approximate real-world data distributions, its limited dataset size restricts these networks from fully exploiting the vast amount of unlabeled data available worldwide \cite{bao2021beit}. Unfortunately, designing encoders based on VFMs, pre-trained in a self/un-supervised manner on extensive unlabeled data, for heterogeneous feature extraction remains largely unexplored in the domain of RGB-T scene parsing. 

Additionally, while symmetric duplex encoders \cite{zhao2023mitigating, deng2021feanet, lv2024context} are prevalently used, they may not be the optimal architectures for heterogeneous feature extraction. This limitation arises from the substantial inherent modality differences \cite{liu2022revisiting} between the RGB and thermal images, as well as the unique advantages provided by vision Transformers (ViTs) and convolutional neural networks (CNNs). RGB images, rich in global semantic cues \cite{seichter2021efficient}, are well-suited for learning through ViTs, while both RGB and thermal images provide gradient-related details (local semantics) \cite{zhou2021gmnet}, which can be more effectively learned through CNNs. Therefore, developing a novel asymmetric duplex encoder, with ViT and CNN employed separately in each branch, for RGB-T scene parsing, stands as an underexplored area of research deserving greater attention.

Moreover, the heterogeneous feature fusion strategy is of considerable importance in RGB-T scene parsing \cite{zhang2019rgb}. The simplistic and indiscriminate strategies used in prior arts \cite{sun2019rtfnet, sun2020fuseseg, ha2017mfnet, deng2021feanet} often cause conflicting feature representations and erroneous scene parsing results \cite{feng2024sne, huang2024roadformer+}. Taking MFNet \cite{ha2017mfnet} as an instance, its feature fusion relies solely on the element-wise concatenation of the heterogeneous features extracted from the duplex encoders at their final stage, treating the features from both modalities equally in the decoder's input. Such a hard-coded feature fusion strategy overlooks the inherent differences in RGB-T feature characteristics and their respective reliability \cite{zhou2021gmnet}, resulting in unsatisfactory performance, particularly under poor illumination conditions. Therefore, another motivation for this article is to design an effective strategy for heterogeneous feature fusion, especially when these features are extracted from different modalities using an asymmetric duplex encoder.

\subsection{Contributions}

\clred{To address the aforementioned limitations, \clredd{we propose the novel integration} of VFMs into RGB-T scene parsing, thus unleashing the representational power derived from self-supervised pre-training on extensive unlabeled data. Given the inherent differences between RGB and thermal images, we first develop a cross-modal spatial prior descriptor (CSPD) based on a CNN to capture robust local spatial patterns from both RGB and thermal images. We then combine the CSPD with a VFM to construct a hybrid, asymmetric encoder, which fuses the heterogeneous features (global context extracted from RGB images and spatial prior captured from RGB-T data) using a progressive heterogeneous feature integrator (PHFI).} To achieve this goal, two key components based on multi-scale deformable attention \cite{zhu2020deformable} are employed within each encoding stage in PHFIs. These components enable the effective integration of multi-scale spatial priors with single-scale global context.
This dual-path, progressive strategy enables a deeper fusion of heterogeneous features. \clr{Additionally, inspired by the deep supervision approaches widely utilized in single-modal scene parsing tasks \cite{sun2024remax}, we introduce an auxiliary task to implicitly enrich the local semantics of fine-grained fused features. This approach effectively reduces noisy and semantically irrelevant features, thereby further improving the overall performance of RGB-T scene parsing.} Our proposed \uline{\textbf{H}ybrid, \textbf{A}symmetric, and \textbf{P}rogressive \textbf{Net}work (\textbf{HAPNet})} outperforms all existing state-of-the-art (SoTA) RGB-T scene parsing methods that primarily leverage symmetric architectures. Specifically, HAPNet achieves improvements of 0.1\%, 1.0\%, and 2.4\% in mean intersection over union (mIoU) on the MFNet \cite{ha2017mfnet}, PST900 \cite{shivakumar2020pst900}, and KP Day-Night \cite{kim2021ms} datasets, respectively. Furthermore, experiments on the NYU-Depth V2 \cite{silberman2012indoor} dataset demonstrate HAPNet's generalizability for RGB-HHA scene parsing. These results demonstrate that HAPNet is particularly effective in handling a wide range of challenging scenarios, especially those characterized by poor illumination conditions or cluttered backgrounds.

In a nutshell, this study makes the following contributions:
\begin{itemize}
    \item We investigate the application of VFMs for RGB-T scene parsing, demonstrating that with appropriate strategies, methods in this domain can significantly benefit from VFMs.
    \item By revisiting the inherent characteristics of RGB and thermal modalities, we propose a hybrid, asymmetric encoder architecture as well as an asymmetric input design, which can fully exploit the complementary strengths of both modalities.
    \item We propose an auxiliary task to further enrich the local semantics of fine-grained fused features. 
    \item Our proposed HAPNet achieves SoTA performance on three publicly available RGB-T scene parsing datasets, and demonstrates promising generalizability for RGB-HHA scene parsing.
\end{itemize}

\subsection{Outline}
The remainder of this article is structured as follows: 
Sect. \ref{sect.related_works} reviews related works.
Sect. \ref{sect.methodology} details our proposed HAPNet.
Sect. \ref{sect.experiments} compares HAPNet with other SoTA methods and presents the ablation study results.
Finally, Sect. \ref{sect.conclusion} concludes this article and discusses possible future work.

\section{Related Work}
\label{sect.related_works}

\begin{figure*}[!t]
    \includegraphics[width=1.0\textwidth]{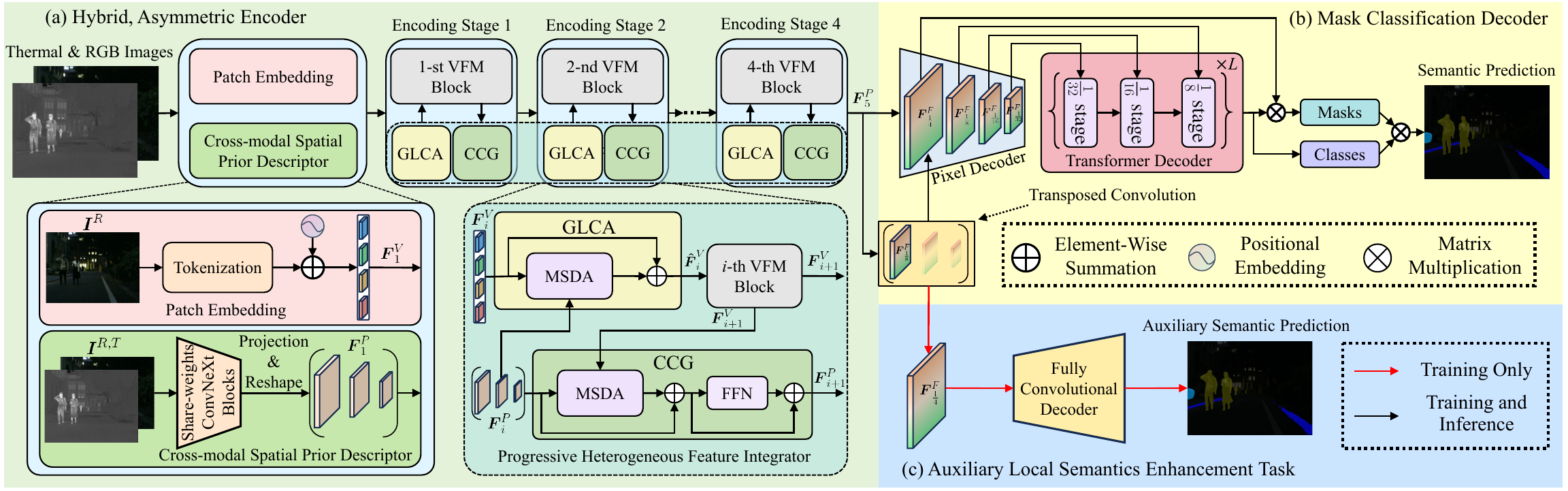}
        \caption{An overview of our proposed HAPNet architecture.}
    \label{fig.architecture}
\end{figure*}
\subsection{Single-Modal Scene Parsing}
\label{intro.single_modal}

The advent of the fully convolutional network (FCN) \cite{long2015fully} catalyzed a wave of research on CNN-based scene parsing networks, which typically utilize convolutions for pixel-wise classification and explore various modules and architectures to enhance overall performance. For instance, the DeepLab series \cite{chen2017deeplab, chen2018encoder} leverages the atrous spatial pyramid pooling module to capture multi-scale context information, while the feature pyramid network (FPN) \cite{lin2017feature} effectively fuses multi-scale features in a fully convolutional fashion for robust dense prediction. Nonetheless, CNNs suffer from inherent limitations, \egi, constrained receptive fields, which impede their effectiveness in comprehensively modeling global context.

Transformers \cite{vaswani2017attention}, originally proposed for natural language processing, have revolutionized computer vision due to their capability to model long-range dependencies. As a representative example, segmentation Transformer (SETR) \cite{zheng2021rethinking} introduces a ViT-based encoder for scene parsing, enabling long-range dependency modeling via self-attention mechanisms. Another approach, SegFormer \cite{xie2021segformer}, employs a hierarchical encoder to achieve multi-scale feature encoding within the Transformer framework, outperforming SETR in handling objects of varying sizes. Additionally, drawing upon earlier works \cite{dai2015convolutional, hariharan2014simultaneous}, the MaskFormer \cite{cheng2021per, cheng2022masked} series formulates scene parsing as a mask classification problem. These networks utilize a Transformer decoder to refine object queries and generate pixel-level predictions by decoding these queries into a set of masks. Given the impressive performance demonstrated by the MaskFormer architecture in dense prediction tasks, we adopt a similar network design for scene parsing in this study.

\subsection{Data-Fusion Scene Parsing}
\label{intro.data_fusion}

\clr{The aforementioned single-modal networks often face challenges in handling complex scenes, especially under poor illumination and adverse weather conditions \cite{zhang2019rgb, zhou2023dbcnet}.} This is primarily because they rely solely on color and texture information present in RGB images \cite{zhou2021gmnet}. Therefore, researchers have explored multi-modal/source data-fusion techniques (with duplex encoders) to overcome these limitations. 
MFNet \cite{ha2017mfnet} pioneers the use of duplex encoders for RGB-T scene parsing. It extracts heterogeneous features using two lightweight backbones and realizes feature fusion via element-wise concatenation, achieving a balance between speed and accuracy. Most subsequent studies generally employ element-wise summation to achieve heterogeneous feature fusion and focus more on encoder and decoder designs for improved performance. For example, RTFNet \cite{sun2019rtfnet} introduces skip connections in the decoder to preserve clear boundaries and fine details. Similarly, FuseSeg \cite{sun2020fuseseg} replaces the ResNet \cite{he2016deep} backbone in RTFNet with the stronger DenseNet \cite{huang2017densely}, enabling richer multi-scale feature extraction. 
\clred{Despite their improved performance, these networks employ simplistic heterogeneous feature fusion strategies that limit their ability to fully exploit the complementary information in RGB-T data.}

To address this drawback, recent works \cite{deng2021feanet, zhou2021gmnet, zhang2023cmx, lv2024context, zhang2021abmdrnet, zhao2023mitigating, shin2023complementary} have resorted to more advanced, learnable feature fusion strategies. For instance, FEANet \cite{deng2021feanet} and GMNet \cite{zhou2021gmnet} introduce attention-based modules to dynamically select important heterogeneous features. \clr{The subsequent research endeavor CMX \cite{zhang2023cmx}, building upon the SegFormer \cite{xie2021segformer} architecture, employs average and max pooling operations along with the attention modules to recalibrate heterogeneous features at the spatial and channel dimensions. Afterwards, its subsequent version CMNeXt \cite{zhang2023delivering} utilizes similar modules based on CNN and pooling operations for feature fusion, and further expands the architecture of CMX to accommodate an arbitrary number of modalities.} Similarly, CAINet \cite{lv2024context} employs different attention modules to model global context and aggregate local semantics across various spatial scales of RGB-T feature maps. Moreover, the ABMDRNet series \cite{zhang2021abmdrnet, zhao2023mitigating} introduces novel modality-wise bridging-then-fusing frameworks, capable of reconstructing pseudo images of one modality using the features extracted from the other modality. This approach successfully reduces modality discrepancy and generates more discriminative fused features. Additionally, inspired by the masked autoencoder (MAE) \cite{he2022masked}, CRM-RGBTSeg \cite{shin2023complementary} randomly masks fixed-size regions within one modality while simultaneously providing complementary regions from another modality to duplex encoders, thus preventing over-reliance on either modality. \clred{In this article, we employ a novel asymmetric duplex encoder to capture both local spatial priors and global context from the RGB-T data. These heterogeneous features are progressively fused for more robust scene parsing.}

\section{Methodology}
\label{sect.methodology}

This section details HAPNet, a hybrid, asymmetric network capable of progressively fusing heterogeneous features extracted from RGB-T data to perform scene parsing. As illustrated in Fig. \ref{fig.architecture}, HAPNet comprises three key components: 
\begin{itemize}
\item An asymmetric, hybrid encoder that extracts heterogeneous features from RGB-T data using a VFM (based on ViT) and a CSPD. In the meantime, these features are progressively fused via PHFI;
\item A Transformer-based mask classification decoder that leverages the fused heterogeneous features to produce semantic predictions;
\item An auxiliary task to further enhance the local semantics of fused heterogeneous features.
\end{itemize}

\subsection{Overall Workflow}

Unlike previous works that utilize symmetric duplex encoder structures, we innovatively introduce an asymmetric, hybrid architecture (based on both a Transformer-based VFM and a lightweight CNN). Compared to the indiscriminate feature extraction strategies utilized by symmetric duplex architectures, our HAPNet effectively leverages the complementary strengths of both modalities. It extracts rich global semantic cues using a VFM, while learning cross-modal local semantics with a CNN. Furthermore, HAPNet conducts effective global-local semantic integration within our proposed PHFIs, resulting in more distinguishable fused features. Consequently, our decoder can more effectively utilize the global context for accurate classification, while maintaining sensitivity to the local details of small-sized objects.
This process starts by evenly dividing the ViT (with $L$ encoder layers) into four blocks, each of which contains $\frac{L}{4}$ encoder layers, and inserting PHFIs into each block to form four encoding stages. The RGB image $\boldsymbol{I}^{R} \in \mathbb{R}^{H \times W \times 3}$ and its corresponding thermal image $\boldsymbol{I}^{T} \in \mathbb{R}^{H \times W \times 3}$ are first fed into the CSPD, extracting the cross-modal spatial prior $\boldsymbol{F}^{P}_1 \in \mathbb{R}^{(\sum_{i=2}^{4}{\frac{HW}{{S_i}^2})} \times D}$, which serves as one of the inputs to following stages, where $S_i=2^{i+1}$ ($i \in [2,4] \cap \mathbb{Z}$) denotes the corresponding stride number. Afterwards, $\boldsymbol{I}^{R}$ is tokenized to form the context feature $\boldsymbol{F}^{V}_1 \in \mathbb{R}^{\frac{HW}{16^2} \times D}$, which serves as the other input to following stages. $\boldsymbol{F}^{V}_1$ and $\boldsymbol{F}^{P}_1$ are then fused in a dual-path, progressive manner using the two key components of our developed PHFI: (1) the global-local context aggregator (GLCA) and (2) the complementary context generator (CCG). Meanwhile, fused features undergo global context encoding in four ViT blocks. This design enables the RGB features to retain fine-grained local semantics while effectively capturing global context through the powerful VFM. Eventually, the fused multi-scale features $\mathcal{F}^{F} = \{ \boldsymbol{F}^F_\frac{1}{S_j} \in \mathbb{R}^{\frac{H}{S_j} \times \frac{W}{S_j} \times D}\}$, where $S_j=2^{i+1}$ ($j \in [1,4] \cap \mathbb{Z}$) from the four encoding stages are obtained and then fed into the subsequent mask classification decoder to generate semantic predictions $\boldsymbol{M}^{P} \in \mathbb{R}^{H \times W}$, which are also utilized by an auxiliary local semantics enhancement task during model training.
\subsection{Hybrid, Asymmetric Encoder}
\label{sec.vit}

\subsubsection{Encoding Pipeline}

We evenly divide $\boldsymbol{I}^{R}$ into non-overlapping patches (resolution: $16 \times 16$ pixels) and project them into $D$-dimensional tokens to form the RGB context feature $\boldsymbol{F}^{V}_1$.
In the $i$-th encoding stage, the input context features $\boldsymbol{F}^{V}_i$ and spatial prior $\boldsymbol{F}^{P}_i$ are first fused in the GLCA to generate the local semantics-enhanced context feature $\boldsymbol{\hat{F}}^{V}_i \in \mathbb{R}^{\frac{HW}{16^2} \times D}$. Subsequently, $\boldsymbol{\hat{F}}^{V}_i$ undergoes context encoding in the ViT block, generating the output $\boldsymbol{F}^{V}_{i+1}$, which is then fed into the CCG to perform another fusion with $\boldsymbol{F}^{P}_i$. This step complements spatial prior $\boldsymbol{F}^{P}_i$ with updated global context from $\boldsymbol{F}^{V}_{i+1}$ to generate $\boldsymbol{F}^{P}_{i+1}$. 
By repeating such a dual-path feature fusion across four encoding stages, we obtain the fused spatial prior $\boldsymbol{F}^{P}_{5}$, which is then recovered to its original three resolutions, forming \{$\boldsymbol{F}^{F}_{\frac{1}{8}}, \boldsymbol{F}^{F}_{\frac{1}{16}}, \boldsymbol{F}^{F}_{\frac{1}{32}}\}$. A $2 \times 2$ transposed convolution is employed to directly create the $\frac{1}{4}$-scale features $\boldsymbol{F}^{F}_{\frac{1}{4}}$ from $\boldsymbol{F}^{F}_{\frac{1}{8}}$. This design avoids the heavy computational overhead of the attention operations required for creating $\boldsymbol{F}^{F}_{\frac{1}{4}}$. Finally, these fused features across four scales constitute $\mathcal{F}^{F}$, which is compatible with current SoTA scene parsing network architectures \cite{cheng2021per, xiao2018unified}.

Solely employing RGB images as the ViT input stems from our hypothesis that explicitly encoding thermal images using VFM may cause convergence issues and representation shift, as discussed in \cite{yin2023dformer}. Furthermore, thermal data often has limited capability in capturing global context, as objects with similar geometries tend to have similar heat signatures. For example, distinguishing between a soccer ball and a basketball or between a cat and a dog is challenging in thermal images. We validate the effectiveness and superior performance of this hybrid, asymmetric network design through ablation studies in Sect. \ref{sect.ablation_study}. 

\subsubsection{Cross-modal Spatial Prior Descriptor}
\label{sec.CSPD}
Compared to Transformers \cite{li2023uniformer, lv2024context}, CNNs have demonstrated superior performance in capturing local semantics, underscoring their significance in computer vision applications that necessitate clear object boundaries. ConvNeXt \cite{liu2022convnet} \clredd{has shown} exceptional performance in capturing rich, robust visual features in comparison with other hierarchical CNN architectures such as ResNet \cite{he2016deep} and its variants \cite{xie2017aggregated}, as evidenced by recent endeavors using ConvNeXt for multi-modal scene parsing \cite{li2023roadformer, bachmann2022multimae}. Given these advantages, we employ basic ConvNeXt blocks to construct our CSPD. Furthermore, incorporating RGB data as a complementary input enables the CSPD to extract more comprehensive local spatial patterns than using thermal images alone. Consequently, we construct our CSPD using two identical, weight-sharing subnetworks, enabling efficient and effective extraction of cross-modal spatial priors from the RGB-T data.

Specifically, $\boldsymbol{I}^{R}$ and $\boldsymbol{I}^{T}$ are independently fed into a series of weight-sharing ConvNeXt blocks, generating multi-scale features {$\mathcal{P}^{R}=\{ \boldsymbol{P}^R_2, \boldsymbol{P}^R_3, \boldsymbol{P}^R_4$}\} and {$\mathcal{P}^{T}=\{ \boldsymbol{P}^T_2, \boldsymbol{P}^T_3, \boldsymbol{P}^T_4$}\}, respectively, where $\boldsymbol{P}^{R,T}_i \in \mathbb{R}^{\frac{H}{S_i} \times \frac{W}{S_i} \times C_{i}}$ represents the features at $i$-th stage, $C_{i}$ and $S_i=2^{i+1}$ ($i\in[2,4]\cap\mathbb{Z}$) denote the corresponding channel and stride numbers, respectively.
We then combine $\boldsymbol{P}^{R}_i$ and $\boldsymbol{P}^{T}_i$ via element-wise summation, and reduce the results to $D$ channels through $1 \times 1$ convolutions, yielding heterogeneous features {$\mathcal{P}^{H}=\{ \boldsymbol{P}^{H}_2, \boldsymbol{P}^{H}_3, \boldsymbol{P}^{H}_4$}\}, where $\boldsymbol{P}^{H}_i \in \mathbb{R}^{\frac{H}{S_i} \times \frac{W}{S_i} \times D}$. The three-scale features are then flattened and concatenated to form the cross-modal spatial prior $\boldsymbol{F}^{P}_1$ which is used for the following encoding stages. We present detailed ablation studies in Sect. \ref{sect.ablation_study} to analyze the impact of different CSPD construction blocks and data input strategies on the overall performance.
\subsubsection{Progressive Heterogeneous Feature Integrator} 
\label{sec.phfi}

The GLCA and CCG, two attention-based components of our PHFI, are responsible for the dual-path feature fusion process before and after each ViT block, respectively. Given $\boldsymbol{F}^{V}_i$ and $\boldsymbol{F}^{P}_i$ of the $i$-th encoding block ($i\in[1,4]\cap\mathbb{Z}$), we first feed them into the GLCA to obtain the local semantics-enhanced input $\boldsymbol{\hat{F}}^{V}_i$. Specifically, $\boldsymbol{F}^{V}_i$ serves as the query, while $\boldsymbol{F}^{P}_i$ acts as the key and value within the GLCA. This process can be formulated as follows:
\begin{equation}
    \boldsymbol{\hat{F}}^{V}_i 
    = \boldsymbol{F}^{V}_i + \kappa_i\mathrm{MHA}(\mathrm{LN}(\boldsymbol{F}^{V}_i), \mathrm{LN}(\boldsymbol{F}^{P}_i)),
\label{eq:injector}
\end{equation}
where $\mathrm{LN}(\cdot)$ represents the layer normalization operation, $\mathrm{MHA}(\cdot,\cdot)$ represents the multi-head attention operation. Following common practices in attention modules such as \cite{huang2021fapn, chen2022vision, li2023roadformer}, we introduce a learnable coefficient $\kappa_i$ to dynamically adjust the weight of the MHA, enabling flexible fusion of the local semantics-enhanced features into the VFM. After the $i$-th ViT block, the output feature maps $\boldsymbol{F}^{V}_{i+1}$ with enriched global context are yielded. $\boldsymbol{F}^{V}_{i+1}$ is then fed into the CCG to fuse its global context with the spatial prior $\boldsymbol{F}^{P}_{i}$. In this process, $\boldsymbol{F}^{P}_i$ serves as the query, while $\boldsymbol{F}^{V}_{i+1}$ acts as the keys and values to perform MHA within the CCG as follows:
\begin{equation}
    \boldsymbol{\hat{F}}^{P}_{i}
    = \boldsymbol{F}^{P}_{i} + \mathrm{MHA}(\mathrm{LN}(\boldsymbol{F}^{P}_{i}), \mathrm{LN}(\boldsymbol{F}^{V}_{i+1})).
\label{eq:extractor}
\end{equation}
After the MHA operation, a feed-forward network (FFN) is applied to further process the fused features, resulting in the updated spatial prior $\boldsymbol{F}^{P}_{i+1}$, \clredd{which has already} incorporated rich local and global semantics and will be inputted to the next encoding stage. In the above two components, the MHA processes are implemented based on multi-scale deformable attention (MSDA) \cite{zhu2020deformable} to reduce computations. Please refer to the supplemental material for more detailed mathematical derivations of MSDA.

\subsection{Mask Classification Decoder}
\label{sec.mask_classification}
Following \cite{cheng2022masked}, our mask classification decoder consists of a pixel decoder and a Transformer decoder.
The former improves the fused heterogeneous features $\{\boldsymbol{F}^{F}_{\frac{1}{8}}, \boldsymbol{F}^{F}_{\frac{1}{16}}, \boldsymbol{F}^{F}_{\frac{1}{32}}\}$, producing $\mathcal{\hat{F}}^P = \{ \boldsymbol{\hat{F}}^F_{\frac{1}{8}}, \boldsymbol{\hat{F}}^F_{\frac{1}{16}},\boldsymbol{\hat{F}}^F_{\frac{1}{32}}\}$, and generates the pixel embedding $\boldsymbol{E}^{P} \in \mathbb{R}^{C \times \frac{H}{4} \times \frac{W}{4}}$ from $\boldsymbol{F}^{F}_\frac{1}{4}$; The latter employs $\mathcal{\hat{F}}^P$ to refine a fixed size of queries, thus producing the mask embedding $\boldsymbol{E}^{M} \in \mathbb{R}^{Q \times C}$, where $Q$ and $C$ represent the numbers of object queries and feature channels, respectively. $\boldsymbol{E}^{M}$ is multiplied by $\boldsymbol{E}^{P}$ to yield mask predictions $\boldsymbol{M}^{M} \in \mathbb{R}^{Q \times \frac{H}{4} \times \frac{W}{4}}$. By resizing $\boldsymbol{M}^{M}$ and associating it with the class predictions $\boldsymbol{E}^{C} \in \mathbb{R}^{Q \times N}$ learned from queries, we obtain the semantic predictions $\boldsymbol{M}^{P}$, where $N$ denotes the number of classes (including a ``no object" class). Readers can refer to \cite{cheng2022masked} for more details on this decoding process.

\subsection{Auxiliary Local Semantics Enhancement Task} 
In our mask classification decoder, $\boldsymbol{E}^{P}$ generated from $\boldsymbol{F}^{F}_\frac{1}{4}$ typically provides rich local (per-pixel) semantics of the scene, while $\boldsymbol{E}^{M}$ generated from the other three scales of features provides information on category-related cluster centers \cite{yang2024polymax}. \clr{To enhance the robustness and discriminability of local semantics utilized by our mask classification decoder, we incorporate deep supervision into $\boldsymbol{F}^{F}_\frac{1}{4}$, effectively reducing noisy and ambiguous patterns within it.} Specifically, we attach $\boldsymbol{F}^{F}_\frac{1}{4}$ with a lightweight fully convolutional decoder head and directly supervise this auxiliary network to produce semantic predictions. After two convolutional layers, the auxiliary semantic predictions with $(N-1)$ channels and the same resolution as $\boldsymbol{F}^{F}_\frac{1}{4}$ are generated (excluding the ``no object" category). This design enables the corresponding encoder layers to produce a $\boldsymbol{F}^{F}_\frac{1}{4}$ with more distinguishable local semantics, further ensuring that the pixel decoder can generate a more distinguishable $\boldsymbol{E}^{P}$, and ultimately leading to improved scene parsing performance without increasing network parameters or computational complexity.

\begin{figure*}[!t]
    \includegraphics[width=0.99\textwidth]{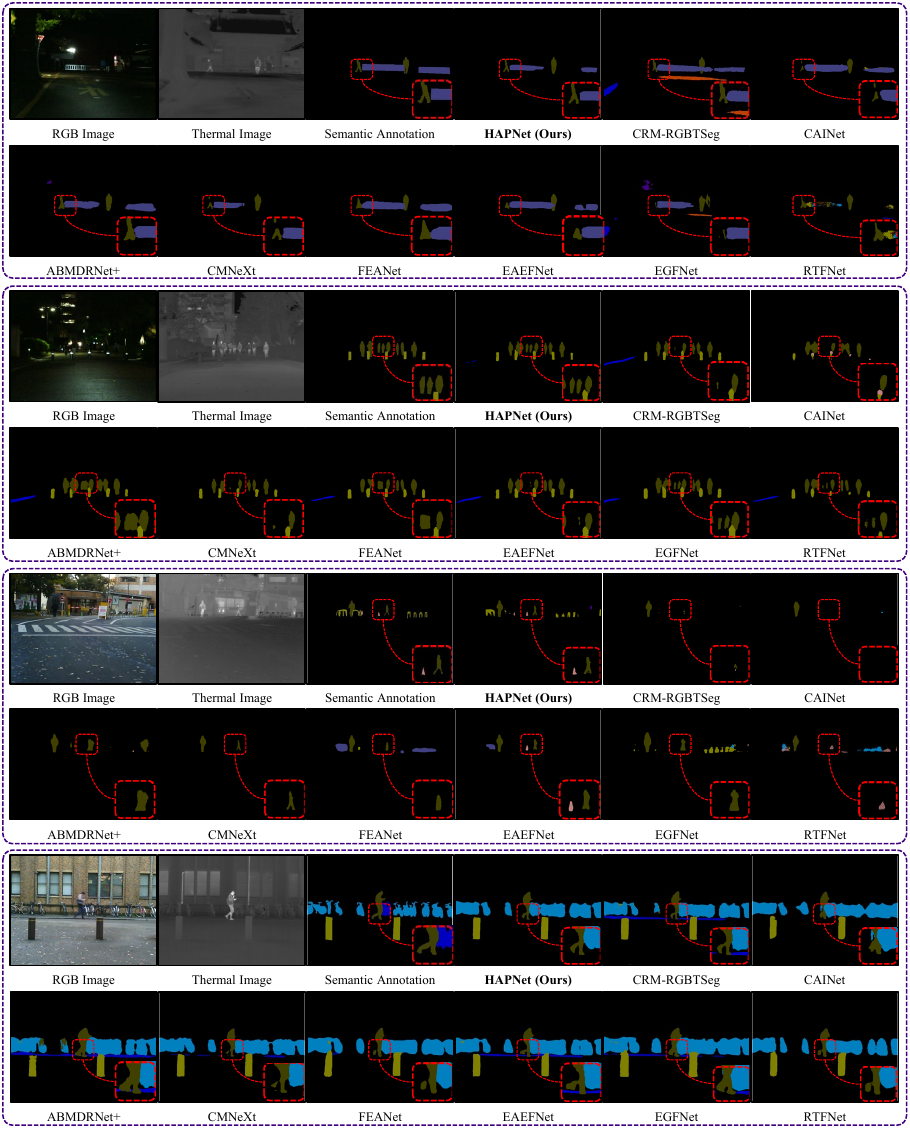}
        \caption{Qualitative comparisons with the SoTA RGB-T scene parsing networks on the MFNet test set, where significantly improved regions are shown with red dashed boxes.}
    \label{fig.mfnet}
\end{figure*}

\subsection{Loss Function}

We employ the cross-entropy (CE) loss $\mathcal{L}_\text{ce}$ for the class prediction of object queries, along with the binary cross-entropy (BCE) loss $\mathcal{L}_\text{bce}$ and dice loss $\mathcal{L}_\text{dice}$ for mask prediction. $\mathcal{L}_\text{ce}$ can be formulated as follows:
\begin{equation}
    \mathcal{L}_{\text{ce}}=-\frac{1}{N} \sum_{i=1}^N \boldsymbol{y}_{c}^i \log (\boldsymbol{p}_{c}^i),
\label{eq:ce}
\end{equation}
where $N$ denotes the total number of object queries, $\boldsymbol{y}_{c}^i$ denotes the one-hot encoded ground-truth vector for $c$ categories, and $\boldsymbol{p}_{c}^i$ stores values indicating the prediction probabilities over $c$ categories. $\mathcal{L}_\text{bce}$ and $\mathcal{L}_\text{dice}$ for mask classification are formulated as follows:
\begin{equation}
    \mathcal{L}_{\text{bce}}=-\frac{1}{M} \sum_{i=1}^M (y_{b}^i \log (p_{b}^i)+(1-y_{b}^i) \log (1-p_{b}^i)),
\label{eq:bce}
\end{equation}
\begin{equation}
    \mathcal{L}_{\text{dice}}=1 - \frac{2\sum_{i=1}^M y_{b}^{i} p_{b}^i}{\sum_{i=1}^M y_{b}^{i} + \sum_{i=1}^M p_{b}^{i}},
\label{eq:dice}
\end{equation}
where $M$ denotes the total number of pixels, $y_{b}^i \in \{0,1\}$ denotes the binary ground truth of one foreground pixel, and $p_{b}^i \in [0,1]$ is a scalar indicating the corresponding prediction probability.
For our local semantics enhancement task, we directly calculate another CE loss $\mathcal{L}_\text{aux}$ over its semantic predictions. The overall loss function can be expressed as follows:
\begin{equation}
\mathcal{L} = \lambda_\text{bce}\mathcal{L}_\text{bce} + \lambda_\text{dice}\mathcal{L}_\text{dice} + \lambda_\text{ce}\mathcal{L}_\text{ce} + \lambda_\text{aux}\mathcal{L}_\text{aux}.
\label{eq.loss}
\end{equation}
We follow \cite{cheng2022masked} and \cite{chen2018encoder} to set $\lambda_\text{bce} = 5.0$, $\lambda_\text{dice} = 5.0$, $\lambda_\text{ce} = 2.0$, and $\lambda_\text{aux} = 0.4$, respectively, and we set $\lambda_\text{ce} = 0.1$ for the ``no object" category. Additionally, we employ bipartite matching \cite{carion2020end} to determine the least-cost assignment during model training.

\begin{table*}[!t] 
	\centering
	\caption{
		Quantitative comparisons (\%) with the SoTA RGB-T scene parsing methods on the MFNet test set. The symbol `-' denotes missing data in the original publication, and the best results are presented in bold font. The values of Acc and IoU for the ``background" category are omitted in the table, but they are still included in the calculation of the corresponding mean values.}
	
	\label{tab.mfnet}
        \fontsize{8}{14}\selectfont
	\setlength{\tabcolsep}{5pt}
	\begin{tabular}{l|cccccccccccccccc|cc}
		\toprule
		\multirow{2}{*}{Methods} & \multicolumn{2}{c}{Car} & \multicolumn{2}{c}{Person} & \multicolumn{2}{c}{Bike} & \multicolumn{2}{c}{Curve} & \multicolumn{2}{c}{Car Stop} & \multicolumn{2}{c}{Guardrail} & \multicolumn{2}{c}{Color Cone} & \multicolumn{2}{c|}{Bump} & \multirow{2}{*}{mAcc} & \multirow{2}{*}{mIoU} \\
		\cline{2-17}
		& Acc & IoU & Acc & IoU & Acc & IoU & Acc & IoU & Acc & IoU & Acc & IoU & Acc & IoU & Acc & IoU \\
		\hline
		\hline
		MFNet ~\cite{ha2017mfnet} & 77.2 & 65.9 & 67.0 & 58.9 & 53.9 & 42.9 & 36.2 & 29.9 & 19.1 & 9.9 & 0.1 & 8.5 & 30.3 & 25.2 & 30.0 & 27.7 & 45.1 & 39.7 \\
		RTFNet ~\cite{sun2019rtfnet} & 93.0 & 87.4 & 79.3 & 70.3 & 76.8 & 62.7 & 60.7 & 45.3 & 38.5 & 29.8 & 0.0 & 0.0 & 45.5 & 29.1 & 74.7 & 55.7 & 63.1 & 53.2 \\
		FuseSeg ~\cite{sun2020fuseseg} & 93.1 & 87.9 & 81.4 & 71.7 & 78.5 & 64.6 & 68.4 & 44.8 & 29.1 & 22.7 & 63.7 & 6.4 & 55.8 & 46.9 & 66.4 & 47.9 & 70.6 & 54.5 \\
		EGFNet ~\cite{zhou2022edge} & \textbf{95.8} & 87.6 & 89.0 & 69.8 & 80.6 & 58.8 & 71.5 & 42.8 & 48.7 & 33.8 & 33.6 & 7.0 & 65.3 & 48.3 & 71.1 & 47.1 & 72.7 & 54.8 \\
		ABMDRNet ~\cite{zhang2021abmdrnet} & 94.3 & 84.8 & 90.0 & 69.6 & 75.7 & 60.3 & 64.0 & 45.1 & 44.1 & 33.1 & 31.0 & 5.1 & 61.7 & 47.4 & 66.2 & 50.0 & 69.5 & 54.8 \\
		ECGFNet ~\cite{zhou2023embedded} & 89.4 & 83.5 & 85.2 & 72.1 & 72.9 & 61.6 & 62.8 & 40.5 & 44.8 & 30.8 & 45.2 & 11.1 & 57.2 & 49.7 & 65.1 & 50.9 & 69.1 & 55.3 \\
		FEANet ~\cite{deng2021feanet} & 93.3 & 87.8 & 82.7 & 71.1 & 76.7 & 61.1 & 65.5 & 46.5 & 26.6 & 22.1 & \textbf{70.8} & 6.6 & \textbf{66.6} & 55.3 & 77.3 & 48.9 & 73.2 & 55.3 \\
		\clr{SFAF-MA} ~\cite{he2023sfaf} & \clr{94.0} & \clr{88.1} & \clr{82.5} & \clr{73.0} & \clr{73.9} & \clr{61.3} & \clr{63.6} & \clr{45.6} & \clr{37.5} & \clr{29.5} & \clr{42.2} & \clr{5.5} & \clr{57.9} & \clr{45.7} & \clr{74.4} & \clr{53.8} & \clr{69.6} & \clr{55.5} \\
		ABMDRNet+ ~\cite{zhao2023mitigating} & 95.2 & 87.1 & \textbf{92.5} & 69.8 & 76.2 & 60.9 & \textbf{72.0} & 47.8 & 42.3 & 34.2 & 66.8 & 8.2 & 64.8 & 50.2 & 63.5 & 55.0 & 74.7 & 56.8 \\
		\clr{LLE-Seg ~\cite{guo2025low}} & \clr{91.8} & \clr{88.6} & \clr{81.3} & \clr{73.2} & \clr{73.7} & \clr{64.8} & \clr{62.5} & \clr{46.8} & \clr{33.7} & \clr{30.0} & \clr{49.1} & \clr{8.8} & \clr{55.7} & \clr{52.5} & \clr{72.9} & \clr{\textbf{62.4}} & \clr{71.6} & \clr{58.4} \\
		CAINet ~\cite{lv2024context} & 93.0 & 88.5 & 74.6 & 66.3 & \textbf{85.2} & \textbf{68.7} & 65.9 & \textbf{55.4} & 34.7 & 31.5 & 65.6 & 9.0 & 55.6 & 48.9 & \textbf{85.0} & 60.7 & 73.2 & 58.6 \\
		EAEFNet ~\cite{liang2023explicit} & 95.4 & 87.6 & 85.2 & 72.6 & 79.9 & 63.8 & 70.6 & 48.6 & 47.9 & 35.0 & 62.8 & 14.2 & 62.7 & 52.4 & 71.9 & 58.3 & \textbf{75.1} & 58.9 \\
		CMX ~\cite{zhang2023cmx} & - & 90.1 & - & 75.2 & - & 64.5 & - & 50.2 & - & 35.3 & - & 8.5 & - & 54.2 & - & 60.6 & - & 59.7 \\
		CMNeXt ~\cite{zhang2023delivering} & - & - & - & - & - & - & - & - & - & - & - & - & - & - & - & - &  & 59.9 \\
		CRM-RGBTSeg ~\cite{shin2023complementary} & - & 90.0 & - & 75.1 & - & 67.0 & - & 45.2 & - & \textbf{49.7} & - & \textbf{18.4} & - & 54.2 & - & 54.4 & - & 61.4 \\
		\hline
		\textbf{HAPNet (Ours)} & 95.1 & \textbf{90.6} & 85.5 & \textbf{75.4} & 79.0 & 67.2 & 67.9 & 51.1 & \textbf{59.5} & 48.4 & 16.0 & 4.2 & 65.0 & \textbf{59.1} & 80.3 & 59.1 & {70.3} & {\color{red}\textbf{61.5}} \\
		\bottomrule
	\end{tabular}
\end{table*}
\begin{figure*}[!t]
    \includegraphics[width=0.99\textwidth]{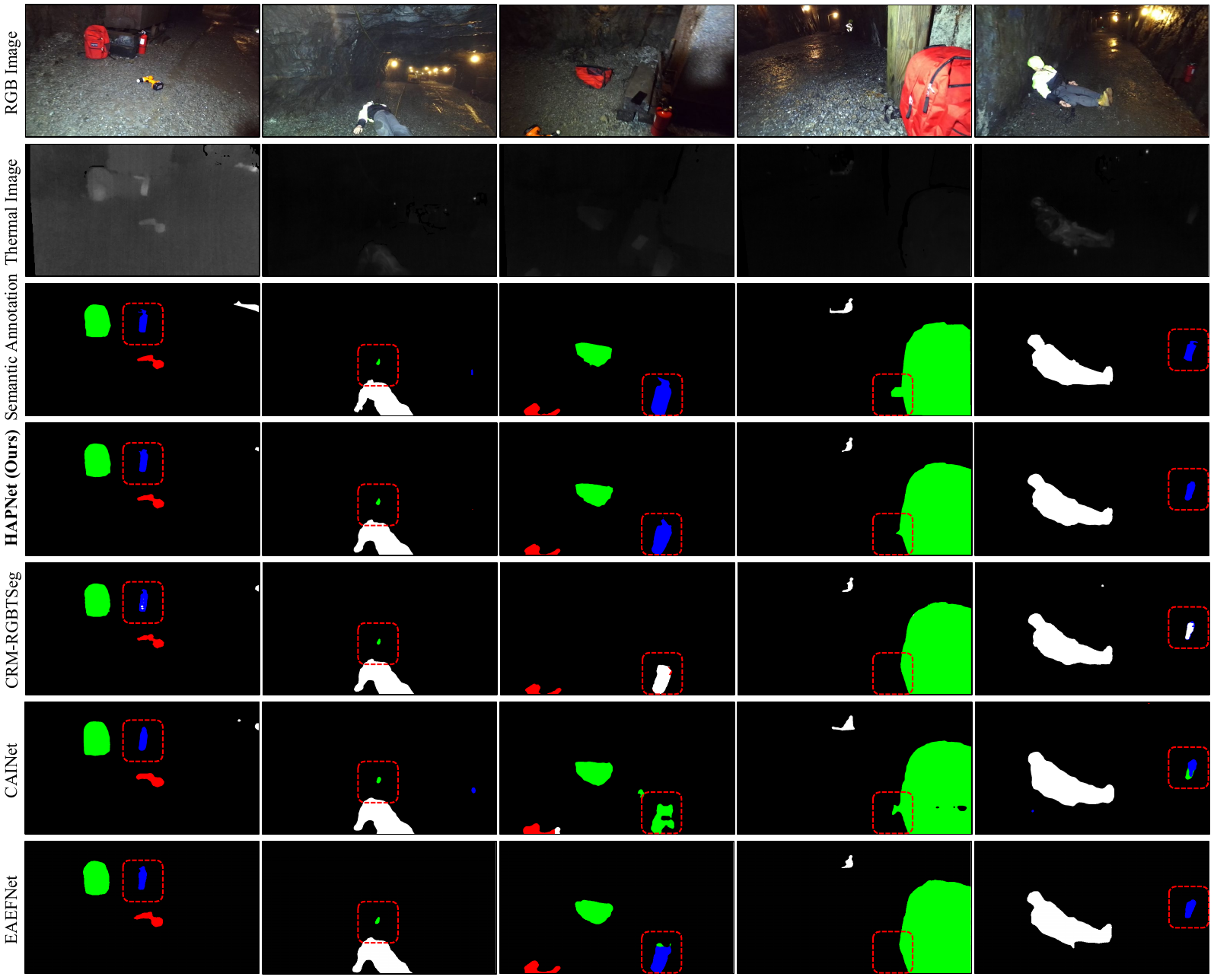}
        \caption{Qualitative comparisons with the SoTA RGB-T scene parsing networks on the PST900 test set, where significantly improved regions are shown with red dashed boxes.}
    \label{fig.pst900}
\end{figure*}
\section{Experiments}
\label{sect.experiments}

We conduct extensive experiments in this article to evaluate the performance of our developed HAPNet both quantitatively and qualitatively. We further quantify the generalizability of HAPNet for RGB-depth/HHA (RGB-D/HHA) scene parsing. The following subsections provide details on the datasets and experimental setup, the evaluation metrics, as well as the comprehensive evaluation of our proposed method.

\subsection{Datasets and Experimental Setup}
\label{sect.dataset_setup}
We first compare HAPNet with other SoTA RGB-T scene parsing networks on the following three public datasets:

\subsubsection{\textbf{MFNet} \cite{ha2017mfnet}}
This is an urban driving scene parsing dataset. An InfReC R500 camera was utilized to capture $1,569$ pairs of synchronized RGB and thermal images at a resolution of $640 \times 480$ pixels.
The dataset provides semantic annotations of nine classes: bike, person, car, road lanes, guardrail, car stop, bump, color cone, and background. We adopt the same dataset splitting strategy as introduced in \cite{ha2017mfnet}.

\subsubsection{\textbf{PST900} \cite{shivakumar2020pst900}}

\begin{figure*}[!t]
    \includegraphics[width=0.99\textwidth]{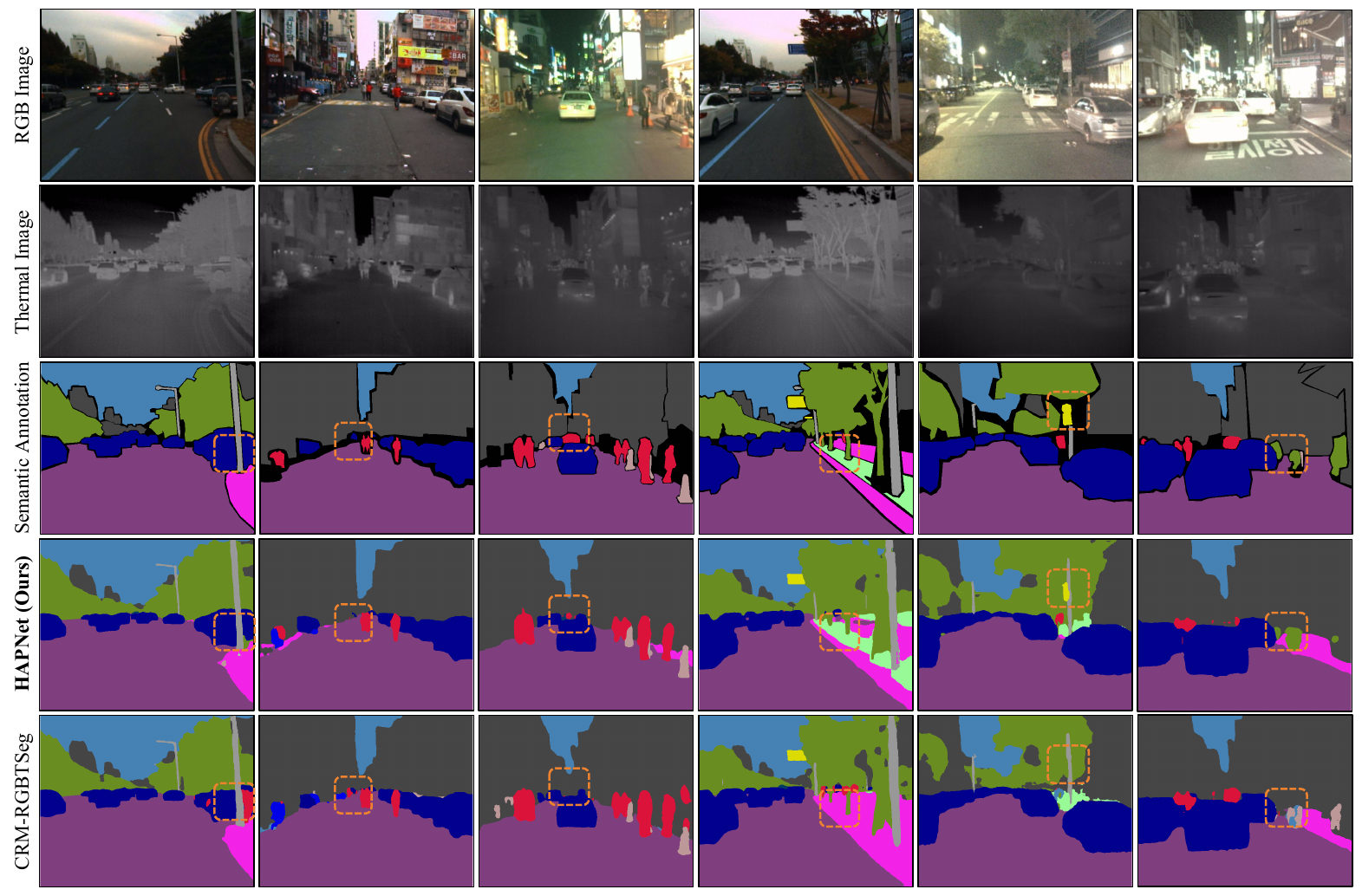}
        \caption{Qualitative comparisons with the SoTA RGB-T scene parsing networks on the KP Day-Night test set, where significantly improved regions are shown with orange dashed boxes.}
    \label{fig.kp}
\end{figure*}

\begin{table*}[!t]
  \centering
  \caption{Quantitative comparisons (\%) with the SoTA RGB-T scene parsing methods on the PST900 test set. The symbol ``-" denotes missing data in the original publication, and the best results are presented in bold font. }
  \label{tab.pst900}
  \fontsize{8}{14}\selectfont
	\setlength{\tabcolsep}{5pt}
  \begin{tabular}{l|cccccccccc|cc}
  \toprule
    \multirow{2}{*}{Methods} & \multicolumn{2}{c}{Background} & \multicolumn{2}{c}{Fire-Extinguisher} & \multicolumn{2}{c}{Backpack} & \multicolumn{2}{c}{Hand-Drill} & \multicolumn{2}{c|}{Survivor} & \multirow{2}{*}{mAcc} & \multirow{2}{*}{mIoU} \\
    \cline{2-11}
    & Acc & IoU & Acc & IoU & Acc & IoU & Acc & IoU & Acc & IoU \\
    \hline
    \hline
    MFNet \cite{ha2017mfnet} & - & 98.6 & - & 41.1 & - & 64.2 & - & 60.3 & - & 20.7 & - & 57.0 \\
    RTFNet \cite{sun2019rtfnet} & - & 98.9 & - & 36.4 & - & 75.3 & - & 52.0 & - & 25.3 & - & 57.6 \\
    EGFNet \cite{zhou2022edge} & - & 99.2 & - & 74.3 & - & 83.0 & - & 71.2 & - & 64.6 & - & 78.5 \\
    ABMDRNet \cite{zhang2021abmdrnet} & - & 98.7 & - & 24.1 & - & 72.9 & - & 54.9 & - & 57.6 & - & 67.3 \\
    FEANet \cite{deng2021feanet} & - & - & - & - & - & - & - & - & - & - & 91.4 & 85.5 \\
    DBCNet \cite{zhou2023dbcnet} & - & 98.9 & - & 62.3 & - & 71.1 & - & 52.4 & - & 40.6 & - & 74.5 \\
    CAINet \cite{lv2024context} & 99.6 & 99.5 & \textbf{95.8} & 80.3 & \textbf{96.0} & 88.0 & 88.3 & 77.2 & 91.3 & 78.6 & \textbf{94.2} & 84.7 \\
    EAEFNet \cite{liang2023explicit} & \textbf{99.8} & 99.5 & 92.2 & 80.4 & 91.0 & 87.7 & 93.0 & 83.9 & 79.3 & 75.6 & 91.1 & 85.4 \\
    GMNet \cite{zhou2021gmnet} & \textbf{99.8} & 99.4 & 90.2 & \textbf{85.1} & 89.0 & 83.8 & 88.2 & 73.7 & 80.8 & 78.3 & 89.6 & 84.1 \\
    CRM-RGBTSeg \cite{shin2023complementary} & - & \textbf{99.6} & - & 79.5 & - & 89.6 & - & 89.0 & - & 82.2 & - & 88.0 \\
    \hline
    \textbf{HAPNet (Ours)} & \textbf{99.8} & \textbf{99.6} & 93.9 & 81.3 & 95.1 & \textbf{92.0} & \textbf{95.5} & \textbf{89.3} & \textbf{85.6} & \textbf{82.4} & 94.0 & {\color{red}\textbf{89.0}} \\
    \bottomrule
  \end{tabular}
\end{table*}

\begin{table*}[t!]
  \centering
  \caption{Quantitative comparisons (\%) with the SoTA RGB-T scene parsing methods on the KP Day-Night test set. The best results are presented in bold font. Although results for Acc and IoU for certain categories are omitted in the table, they are still included in the calculations of the corresponding mean values.}
  \label{tab.kp}
  \fontsize{8}{15}\selectfont
	\setlength{\tabcolsep}{5pt}
  \begin{tabular}{l|ccccccccccccccccc|c}
  \toprule
    Methods & \rotatebox[origin=c]{90}{Road} & \rotatebox[origin=c]{90}{Sidewalk} & \rotatebox[origin=c]{90}{Building} & \rotatebox[origin=c]{90}{Fence} & \rotatebox[origin=c]{90}{Pole} & \rotatebox[origin=c]{90}{Traffic light} & \rotatebox[origin=c]{90}{Traffic sign} & \rotatebox[origin=c]{90}{Vegetation} & \rotatebox[origin=c]{90}{Terrain} & \rotatebox[origin=c]{90}{Sky} & \rotatebox[origin=c]{90}{Person} & \rotatebox[origin=c]{90}{Rider} & \rotatebox[origin=c]{90}{Car} & \rotatebox[origin=c]{90}{Truck} & \rotatebox[origin=c]{90}{Bus} & \rotatebox[origin=c]{90}{Motorcycle} & \rotatebox[origin=c]{90}{Bicycle} & mIoU \\
    \hline
    \hline
    MFNet \cite{ha2017mfnet} & 93.5 & 23.6 & 75.1 & 0.1 & 9.1 & 0.0 & 0.0 & 69.3 & 0.2 & 90.4 & 24.0 & 0.0 & 69.6 & \textbf{0.3} & 0.3 & 0.0 & 0.6 & 24.0 \\
    RTFNet \cite{sun2019rtfnet} & 94.6 & 39.4 & 86.6 & 0.6 & 0.0 & 0.0 & 0.0 & 81.7 & 3.7 & 92.8 & 58.4 & 0.0 & 87.7 & 0.0 & 0.0 & 0.0 & 0.5 & 28.7 \\
    CMX \cite{zhang2023cmx} & 97.7 & 53.8 & 90.2- & 47.1 & 46.2 & 10.9 & 45.1 & 87.2 & 34.3 & 93.5 & 74.5 & 0.0 & 91.6 & 0.0 & 59.7 & 46.1 & 0.2 & 46.2 \\
    CRM-RGBTSeg \cite{shin2023complementary} & \textbf{99.0} & \textbf{61.9} & \textbf{91.8} & \textbf{58.7} & 50.6 & 39.2 & 55.3 & \textbf{89.2} & 23.2 & 94.3 & \textbf{85.2} & 2.9 & \textbf{95.3} & 0.0 & \textbf{80.5} & \textbf{66.2} & \textbf{54.6} & 55.2 \\
    \hline
    \textbf{HAPNet (Ours)} & 98.6 & 59.3 & 91.5 & 57.0 & \textbf{58.0} & \textbf{41.0} & \textbf{56.8} & 87.8 & \textbf{30.4} & \textbf{94.8} & 81.8 & \textbf{21.3} & 94.2 & 0.0 & 69.7 & 49.9 & 43.8 & {\color{red}\textbf{57.6}} \\
    \bottomrule
  \end{tabular}
\end{table*}

\begin{table}[!htbp]
  \centering
  \caption{Quantitative comparisons (\%) with the SoTA data-fusion scene parsing networks on the NYU-Depth V2 test set. The symbol ``-" denotes missing data in the original publication, and the best results are presented in bold font.}
  \label{tab.nyu}
  \fontsize{8}{15}\selectfont
	\setlength{\tabcolsep}{5pt}
  \begin{tabular}{l|l|ccccccccccccccccc|c}
  \toprule
    Input Data & Methods & Pixel Acc & mAcc & mIoU & Rank \\
    \hline
    \hline
    \multirow{7}{*}{RGB-D} & OmniVec \cite{srivastava2024omnivec} & - & - & {\color{red}\textbf{60.8}} & {\color{red}\textbf{1}} \\
    & Omnivore \cite{girdhar2022omnivore} &  &  & 56.8 & 8 \\
    & TokenFusion \cite{wang2022multimodal} & 79.0 & 66.9 & 54.2 & 16 \\
    & DFormer-L \cite{yin2023dformer} & - & - & 57.2 & 5  \\
    & AsymFormer \cite{du2023asymformer} & 78.5 & - & 54.1 & 17  \\
    & RTFNet \cite{sun2019rtfnet} & - & 64.8 & 49.1 & 59 \\
    & ECGFNet \cite{zhou2023embedded} & - & 65.2 & 51.5 & 37 \\
    \hline
    \multirow{4}{*}{RGB-HHA} & $\text{CMX}$ \cite{zhang2023cmx} & - & - & 56.9 & 6  \\
    & $\text{CMNeXt}$ \cite{zhang2023delivering} & - & - & 56.9 & 6 \\
    & $\text{SA-Gate}$ \cite{chen2020bi} & - & - & 52.4 & 31 \\
    & $\text{\textbf{HAPNet}}$ \textbf{(Ours)} & \textbf{79.2} & \textbf{68.8} & {\color{blue}55.0} & {\color{blue}15} \\
    \bottomrule
  \end{tabular}
\end{table}

\begin{table*}[!htbp]
  \centering
  \caption{An ablation study (\%) on different data input strategies on the MFNet test set.}
  \label{tab.input_ablation}
  \fontsize{8}{15}\selectfont
	\setlength{\tabcolsep}{7.7pt}
  \begin{tabular}{l|cc|cccccccc|cc}
  \toprule
    \multirow{2}{*}{Strategies} & \multirow{2}{*}{VFM} &\multirow{2}{*}{CSPD} & Car & Person & Bike & Curve & Car Stop & Guardrail & Color Cone & Bump & \multirow{2}{*}{mAcc} & \multirow{2}{*}{mIoU} \\
    \cline{4-11}
    &&& IoU & IoU  & IoU  & IoU  & IoU & IoU & IoU & IoU \\
    \hline
    \hline
    A & RGB + T & RGB + T & 89.3 & 73.9 & 66.1 & 48.0 & 32.4 & 4.4 & 55.7 & 62.6 & 74.3 & 58.9 \\
    B & RGB + T & RGB & 88.3 & 69.1 & 64.3 & 44.3 & 32.3 & 4.3 & 52.9 & 44.5 & 67.0 & 55.3 \\
    C & RGB + T & Thermal & 85.5 & 72.7 & 54.6 & 39.3 & 28.5 & \textbf{8.2} & 46.1 & 55.4 & 71.5 & 54.2 \\
    \textbf{D (Ours)} & RGB & RGB + T & 89.1 & \textbf{76.0} & \textbf{67.2} & \textbf{50.6} & \textbf{41.3} & 7.9 & \textbf{58.6} & 58.7 & \textbf{75.1} & \textbf{60.9} \\
    \hline
    E & RGB & RGB & 87.9 & 62.5 & 63.8 & 40.7 & 28.2 & 5.6 & 52.1 & 47.1 & 66.7 & 53.9 \\
    F & RGB & Thermal & 86.7 & 73.9 & 60.3 & 39.1 & 33.0 & 7.3 & 53.6 & \textbf{59.0} & 74.6 & 56.7 \\
    \hline
    G & Thermal & RGB + T & \textbf{89.6} & 74.6 & 65.1 & 48.9 & 33.9 & 1.5 & 53.9 & 56.8 & 68.8 & 58.1 \\
    H & Thermal & RGB & 88.4 & 69.3 & 65.3 & 43.2 & 29.7 & 4.4 & 53.4 & 44.4 & 67.4 & 55.1 \\
    I & Thermal & Thermal & 84.5 & 71.8 & 53.8 & 35.0 & 24.3 & 3.8 & 38.4 & 54.5 & 66.6 & 51.5 \\
    \bottomrule
  \end{tabular}
\end{table*}

\begin{table*}[!htbp]
  \centering
  \caption{An ablation study (\%) on different CSPD construction blocks on the MFNet test set.}
  \label{tab.CSPD}
  \fontsize{8}{15}\selectfont
	\setlength{\tabcolsep}{5.7pt}
  \begin{tabular}{l|cccccccc|cc}
  \toprule
    \multirow{2}{*}{Encoders} & Car & Person & Bike & Curve & Car Stop & Guardrail & Color Cone & Bump & \multirow{2}{*}{mAcc} & \multirow{2}{*}{mIoU} \\
    \cline{2-9}
    & IoU & IoU & IoU  & IoU  & IoU & IoU & IoU & IoU \\
    \hline
    \hline
    Duplex ResNet-101 \cite{he2016deep} sharing weights  & 66.3 & 48.0 & 38.4 & 19.9 & 9.1 & 0.0 & 29.6 & 23.6 & 42.3 & 36.8 \\
    Duplex Swin-Transformer-S \cite{liu2021swin} sharing weights & \textbf{89.7} & 74.1 & 67.2 & 47.6 & 31.0 & 6.3 & 52.7 & 57.0 & 74.6 & 58.2 \\
    Duplex MiT-B4 sharing weights \cite{xie2021segformer} sharing weights & 87.8 & 74.7 & 64.9 & 48.4 & 40.9 & \textbf{13.1} & 51.4 & 48.1 & 73.2 & 58.6 \\
    Duplex ConvNeXt-S \cite{liu2022convnet} with separate weights & 89.6 & 74.6 & 66.6 & 46.3 & \textbf{45.7} & 10.2 & 56.6 & \textbf{62.5} & 73.6 & \textbf{61.2} \\
    \textbf{Duplex ConvNeXt-S \cite{liu2022convnet} sharing weights (Ours)} & 89.1 & \textbf{76.0} & \textbf{67.2} & \textbf{50.6} & 41.3 & 7.9 & \textbf{58.6} & 58.7 & \textbf{75.1} & 60.9 \\
    \bottomrule
  \end{tabular}
\end{table*}

\begin{table*}[!htbp]
  \centering
  \caption{An ablation study (\%) on the effectiveness of GLCA and CCG on the MFNet test set. When both components are removed, the spatial priors extracted using CSPD and the global context extracted using ViT are fused via element-wise summation after resolution alignment.}
  \label{tab.modules_ablation}
  \fontsize{8}{15}\selectfont
	\setlength{\tabcolsep}{6.5pt}
  \begin{tabular}{cc|c|cccccccc|cc}
  \toprule
    \multirow{2}{*}{GLCA} & \multirow{2}{*}{CCG} & \multirow{2}{*}{Fusion Strategies} & Car & Person & Bike & Curve & Car Stop & Guardrail & Color Cone & Bump & \multirow{2}{*}{mAcc} & \multirow{2}{*}{mIoU} \\
    \cline{4-11}
   &&& IoU & IoU  & IoU  & IoU  & IoU & IoU & IoU & IoU \\
    \hline
    \hline
      & & {$\text{Element-wise Summation}$} & 89.2 & 73.5 & 65.7 & 50.0 & 33.0 & 4.1 & 52.0 & 55.4 & 71.1 & 57.9 \\
    $\checkmark$ & & Local Semantics Enhanced & \textbf{89.9} & 74.7 & 67.0 & 50.0 & 37.0 & 5.8 & 52.9 & 55.9 & 72.9 & 59.1 \\
     & $\checkmark$ & Global Context Updated & 89.6 & 74.5 & 67.0 & 46.6 & 36.9 & 2.9 & 53.3 & 53.3 & 72.4 & 58.0 \\
    $\checkmark$ & $\checkmark$ & \textbf{Both (Ours)} & 89.1 & \textbf{76.0} & \textbf{67.2} & \textbf{50.6} & \textbf{41.3} & \textbf{7.9} & \textbf{58.6} & \textbf{58.7} & \textbf{75.1} & \textbf{60.9} \\
    \bottomrule
  \end{tabular}
\end{table*}

\begin{table*}[!htbp]
  \centering
  \caption{\clr{An ablation study (\%) on different symmetric and asymmetric encoder architectures on the MFNet test set.}}
  \label{tab.backbone_ablation}
  \fontsize{8}{15}\selectfont
	\setlength{\tabcolsep}{7.3pt}
  \begin{tabular}{l|cccccccc|cc}
  \toprule
    \multirow{2}{*}{Methods (encoder)} & Car & Person & Bike & Curve & Car Stop & Guardrail & Color Cone & Bump & \multirow{2}{*}{mAcc} & \multirow{2}{*}{mIoU} \\
    \cline{2-9}
    & IoU & IoU  & IoU  & IoU  & IoU & IoU & IoU & IoU \\
    \hline
    \hline
    Duplex $\text{BEiTv2-B/16}$ \cite{peng2022beitv2} concatenation & 65.7 & 57.4 & 12.7 & 27.7 & 8.5 & 0.0 & 14.7 & 20.6 & 40.9 & 33.7 \\
    Duplex $\text{BEiTv2-B/16}$ \cite{peng2022beitv2} summation & 65.4 & 57.1 & 23.3 & 28.0 & 8.5 & 0.0 & 19.9 & 16.2 & 42.6 & 35.0 \\
    Duplex ConvNeXt-S \cite{liu2022convnet} concatenation & 89.7 & 74.1 & 66.4 & 45.3 & 29.8 & 4.8 & 52.8 & 55.5 & 69.2 & 57.4 \\
    Duplex ConvNeXt-S \cite{liu2022convnet} summation & \textbf{90.7} & 74.6 & 66.9 & 46.8 & 32.9 & 6.6 & 56.6 & 52.7 & 72.0 & 58.5 \\
    \clr{HAPNet CMNeXt-B4} \cite{zhang2023delivering} & \clr{88.5} & \clr{74.0} & \clr{62.5} & \clr{44.9} & \clr{41.1} & \clr{\textbf{12.0}} & \clr{53.3} & \clr{\textbf{59.0}} & \clr{71.6} & \clr{59.3} \\
    \clr{\textbf{HAPNet (Ours)}} & \clr{89.1} & \clr{\textbf{76.0}} & \clr{\textbf{67.2}} & \clr{\textbf{50.6}} & \clr{\textbf{41.3}} & \clr{7.9} & \clr{\textbf{58.6}} & \clr{58.7} & \clr{\textbf{75.1}} & \clr{\textbf{60.9}} \\
    \bottomrule
  \end{tabular}
\end{table*}

\begin{table*}[!htbp]
  \centering
  \caption{An ablation study (\%) on different VFMs on the MFNet test set. ``MM" represents multi-modal pre-training. BEiT and BEiTv2 are trained via a self-supervised learning strategy named masked image modeling, while DINOv2 is trained via discriminative self-supervised learning strategy.}
  \label{tab.vfm_ablation}
  \fontsize{8}{15}\selectfont
	\setlength{\tabcolsep}{3.2pt}
  \begin{tabular}{lcc|cccccccc|cc}
  \toprule
    \multirow{2}{*}{VFM} & \multirow{2}{*}{Pre-training Strategy} & \multirow{2}{*}{Dataset} & Car & Person & Bike & Curve & Car Stop & Guardrail & Color Cone & Bump & \multirow{2}{*}{mAcc} & \multirow{2}{*}{mIoU} \\
    \cline{4-11}
    &&& IoU & IoU  & IoU  & IoU  & IoU & IoU & IoU & IoU \\
    \hline
    \hline
    DeiT-B/16 \cite{touvron2021training} & Supervised & ImageNet-1K & 90.4 & 75.2 & 66.6 &47.6 & 36.3 & 0.7 & \textbf{59.9} & \textbf{60.0} & 68.3 & 59.4 \\
    AugReg-B/16 \cite{steiner2022train} & Supervised & ImageNet-22K & 90.3 & 74.9 & 66.8 & 48.2 & 35.8 & 6.0 & 56.6 & 56.5 & 74.7 & 59.3 \\
    BEiT-B/16 \cite{bao2021beit} & Self-Supervised & ImageNet-22K & \textbf{90.7} & 75.0 & \textbf{67.6} & 48.6 & 39.1 & 4.2 & 55.0 & 59.0 & 72.5 & 59.7 \\
    DINOv2-B/14 \cite{oquab2023dinov2} & Self-Supervised & LVD-142M (MM) \cite{oquab2023dinov2} & 90.0 & 74.7 & 66.8 &45.1 & \textbf{42.1} & \textbf{9.7} & 56.2 & 59.7 & \textbf{76.4} & 60.3 \\
    BEiTv2-B/16 \cite{peng2022beitv2} & Self-Supervised & ImageNet-22K {$\text{(MM)}$} & 89.1 & \textbf{76.0} & 67.2 & \textbf{50.6} & 41.3 & 7.9 & 58.6 & 58.7 & 75.1 & \textbf{60.9} \\
    \bottomrule
  \end{tabular}
\end{table*}

This dataset contains $894$ pairs of RGB and thermal images captured in challenging underground environments (originally released on the DARPA Subterranean Challenge). The RGB images were acquired using a Stereolabs ZED Mini stereo camera, while the thermal images were captured using a FLIR Boson 320 camera, both at a resolution of $1,280 \times 720$ pixels. The dataset has five categories of semantic annotations: background, fire extinguisher, backpack, hand drill, and survivor. We adopt the same data splitting strategy\footnote{The number of image pairs being published is fewer than what was reported in publication \cite{shivakumar2020pst900}.} as described in \cite{shivakumar2020pst900}. 

\subsubsection{\textbf{KP Day-Night} \cite{kim2021ms}}

This dataset contains $950$ well-rectified RGB-T image pairs (resolution: $640 \times 512$ pixels) captured in urban driving scenes. This study \cite{kim2021ms} provides semantic annotations of $19$ categories (identical to the CityScapes \cite{cordts2016cityscapes} dataset). We adopt the same data splitting strategy presented in \cite{shin2023complementary} to train and evaluate our method.

Furthermore, we conduct an additional experiment on the NYU-Depth V2 dataset \cite{silberman2012indoor} to evaluate the generalizability of our network for RGB-D/HHA scene parsing.

\subsubsection{\textbf{NYU-Depth V2} \cite{silberman2012indoor}}

This dataset has been widely used for the evaluation of indoor \clredd{scene parsing performance}. It provides $1,449$ pairs of RGB and depth images (resolution: $480 \times 640$ pixels), and semantic annotations for $40$ categories. We adopt the same dataset splitting strategy as presented in \cite{yin2023dformer}. 

\subsection{Evaluation Metrics}
We employ two widely used evaluation metrics: accuracy (Acc) and IoU, to quantify the scene parsing performance with respect to each category. The mean values of these two metrics across all categories (denoted as mIoU and mAcc, respectively) are also computed to quantify the comprehensive network performance.

\subsection{Implementation Details}
Our model was trained for 200 epochs on an NVIDIA RTX 3090 GPU, utilizing the AdamW optimizer \cite{loshchilov2018decoupled} with an initial learning rate of $2 \times 10^{-4}$ and a weight decay of $5 \times 10^{-2}$. Following the common practice employed in \cite{bao2021beit}, we apply a layer-wise learning rate decay of $0.9$ for our VFM encoder. For the experiments conducted on the NYU dataset, we employ RGB images and the HHA encoding of depth images as model input. 
We conduct all ablation studies on the widely utilized MFNet dataset, with the auxiliary task omitted. \clredd{By comparing the results in Tables \ref{tab.mfnet} and \ref{tab.backbone_ablation}, we observe that the performance improvement attributed to the auxiliary task is $0.6\%$ in mIoU (calculated as an average across multiple experiments and subject to a fluctuation of approximately $\pm0.2\%$).} These results demonstrate the effectiveness of the auxiliary task.

\subsection{Comparisons with SoTA Scene Parsing Networks}
\label{sect.exp_mfnet}

We conduct quantitative comparisons with $15$, $10$, and four SoTA RGB-T scene parsing networks on the MFNet \cite{ha2017mfnet}, PST900 \cite{shivakumar2020pst900}, and KP Day-Night \cite{kim2021ms} datasets, respectively. The results are detailed in Tables \ref{tab.mfnet}, \ref{tab.pst900}, and \ref{tab.kp}, respectively. Additionally, we present the qualitative comparisons on these three datasets, as shown in Fig. \ref{fig.mfnet}, \ref{fig.pst900}, and \ref{fig.kp}, respectively.

Specifically, our proposed HAPNet achieves the highest mIoU across all datasets, outperforming SoTA methods by $0.1$-$21.8\%$ on the MFNet dataset, $1.0$-$32.0\%$ on the PST900 dataset, and $2.4$-$33.6\%$ on the KP day-night dataset, respectively. These results demonstrate the robustness and effectiveness of HAPNet for RGB-T scene parsing in various scenarios, ranging from complex urban driving scenes to challenging underground scenes. \clred{Notably, despite performance variations on rare categories (e.g., "guardrail") in the MFNet dataset, our approach consistently outperforms CRM-RGBTSeg \cite{shin2023complementary}, a state-of-the-art Transformer-based method using the mask classification paradigm. This demonstrates the efficacy of our hybrid asymmetric architecture for heterogeneous feature fusion.}
Compared to CMNeXt \cite{zhang2023delivering}, a general-purpose multi-modal data-fusion method that also employs an asymmetric encoder architecture, HAPNet achieves an improvement of $1.6\%$ in mIoU on the MFNet. This performance improvement can be attributed to our method's utilization of a more powerful VFM and a more advanced multi-scale cross-attention mechanism that integrates global context from the VFM and multi-scale spatial priors provided by a CNN, in contrast to CMNeXt's reliance on CNN and pooling-based feature fusion strategies. Consequently, HAPNet generates more discriminative fused features for RGB-T scene parsing. The distinct parsing results in Figs. \ref{fig.mfnet}, \ref{fig.pst900}, and \ref{fig.kp} also demonstrate the superior capability of HAPNet for visual perception under poor illumination conditions.

In addition, HAPNet demonstrates promising generalizability for RGB-D/HHA scene parsing. As shown in Table \ref{tab.nyu}, our method outperforms all other RGB-T scene parsing networks on the NYU Depth V2 dataset. Specifically, HAPNet achieves a $3.5\%$ higher mIoU than ECGFNet \cite{zhou2023embedded}, and a $5.9\%$ higher mIoU than RTFNet \cite{sun2019rtfnet}. \clred{However, these results suggest that HAPNet still lags behind SoTA methods, developed specifically for RGB-D/HHA data fusion, including Omnivore \cite{girdhar2022omnivore}, DFormer-L \cite{yin2023dformer}, and OmniVec \cite{srivastava2024omnivec} (mIoU is lower by $2.2$-$5.8\%$).} Additionally, it underperforms CMX and CMNeXt \cite{zhang2023cmx,zhang2023delivering}, two SoTA general-purpose multi-modal scene parsing methods by $1.9\%$ in mIoU. 
\clred{We hypothesize that this performance gap stems from thermal images providing richer complementary features than depth/HHA images. While depth/HHA images primarily convey geometric information, thermal images capture both fine-grained spatial priors and semantic cues through heat signatures (e.g., distinguishing animate from inanimate objects). These characteristics are particularly well-suited to our proposed CSPD and CCG modules. Consequently, these inherent distributional differences between thermal and depth/HHA modalities limit the applicability of our architecture to RGB-D/HHA scene parsing.}

Despite achieving SoTA performance, HAPNet demonstrates competitive real-time performance. \clr{Our network contains $169.1$M parameters and achieves an inference speed of $\sim23$ frames per second (FPS) when processing images at a resolution of $480 \times 480$ pixels on an NVIDIA RTX 4090-D GPU.} This performance nearly meets the real-time processing requirements for autonomous driving systems.

\subsection{Ablation Study}
\label{sect.ablation_study}

We first explore the performance of RGB-T scene parsing with respect to different input data combination strategies. Given the asymmetric two-branch architecture, the RGB-T data can be fed into our HAPNet using nine different strategies to the VFM and CSPD, respectively, as shown in Table \ref{tab.input_ablation}. Among them, strategies E and I can be considered as single-modal versions (only RGB or thermal images are encoded by the network), yielding mIoU scores of only $53.9\%$ and $51.5\%$, respectively. On the other hand, strategies A-D, and F-H, which employ both RGB and thermal images, achieve superior performance, validating the importance of modality complementation. Notably, strategy D, employing RGB images as inputs to the VFM branch while using a combination of RGB-T data in the CSPD, achieves the highest mIoU of $60.9\%$. This result demonstrates that encoding thermal images using the VFM may deteriorate features and degrade the overall performance, which verifies our hypothesis stated in Sect. \ref{sect.intro}.

Our CSPD extracts multi-scale spatial priors from both RGB and thermal images, which are subsequently fused with the VFM context features to implicitly enrich local semantics. To further explore the most suitable structure for the CSPD, we replace the basic ConvNeXt blocks with other mainstream hierarchical encoder structures. As detailed in Table \ref{tab.CSPD}, ConvNeXt demonstrated superior performance, achieving a mIoU of $60.9\%$, outperforming established architectures such as ResNet, Swin-Transformer, and the Mix Vision Transformer (MiT-B4), the latter being a heavy structure previously employed in another RGB-X scene parsing network \cite{zhang2023delivering}. 

In addition, we investigate the impact of employing separate weights for the two modalities, and this strategy yields a slightly better result. However, considering the trade-off between accuracy and model complexity, we still adopt the weight-sharing strategy, as the weight-separating strategy only achieves an improvement of $0.3\%$ in terms of mIoU while incurring a substantial increase in the modal parameters.

Furthermore, we evaluate the effectiveness of GLCA and CCG. As presented in Table \ref{tab.modules_ablation}, removing either of them leads to a significant performance deterioration of HAPNet. The baseline setup, which could only perform cross-modal feature fusion via element-wise summation between spatial priors and global context, exhibits a $3.0\%$ lower mIoU compared to the complete HAPNet.

To validate the effectiveness of our proposed asymmetric architecture, we first construct baseline architectures incorporating symmetric duplex encoders based on BEiTv2 \cite{peng2022beitv2} and ConvNeXt \cite{liu2022convnet}, respectively. As shown in Table \ref{tab.backbone_ablation}, such baseline architectures fail to achieve performance comparable to that of our developed HAPNet. Specifically, the duplex, symmetric ConvNeXt and BEiTv2 architectures fall short of HAPNet, with a decrease in mIoU by $3.5$-$2.4\%$ and $27.2$-$25.9\%$, respectively. \clr{These findings lend further support to our hypothesis stated in Sect. \ref{sect.intro}, suggesting that applying asymmetric architectures to RGB and thermal modalities can play to their own strengths. Importantly, the disappointing performance of the symmetric duplex BEiTv2 encoder suggests that directly applying VFMs for the RGB-T scene parsing task might not be suitable.} \clr{We further compare the encoder of HAPNet with another best-performing asymmetric encoder proposed in \cite{zhang2023delivering}. In order to fairly compare the performance of the two encoder architectures, we combined CMNeXt's encoder with HAPNet's decoder, abbreviated as HAPNet CMNeXt-B4, as shown in Table \ref{tab.backbone_ablation}, and the mIoU exhibits a decrease of $1.6\%$, further underscoring the superiority of our encoder design compared to other architectures for RGB-T scene parsing.}

Moreover, we also evaluate the compatibility of HAPNet with various SoTA VFMs, including those trained through traditional supervised learning (DeiT \cite{touvron2021training} and AugReg \cite{steiner2022train}) as well as the most recent models developed based on self-supervised pre-training (BEiT \cite{bao2021beit}, BEiTv2 \cite{peng2022beitv2}, and DINOv2 \cite{oquab2023dinov2}). As shown in Table \ref{tab.vfm_ablation}, BEiTv2 achieves the highest mIoU of $60.9\%$, outperforming all other models on the RGB-T scene parsing task. Given BEiTv2's superior performance, we empirically determine it to be the most suitable VFM for this task, thereby adopting BEiTv2 as the default VFM within HAPNet.

\section{Conclusion and Future Work}
\label{sect.conclusion}
In this article, we revisited the design of heterogeneous feature extraction and fusion strategies. Drawing upon the inherent characteristics of RGB and thermal data, we developed a novel hybrid, asymmetric network to capitalize on their unique strengths, leveraging the powerful VFM for this specific task. We first developed a cross-modal spatial prior descriptor to capture local semantics jointly from RGB-T data. Additionally, we designed a progressive heterogeneous feature integrator, consisting of a global-local context aggregator and a complementary context generator, to fuse heterogeneous features more effectively. We also introduced an auxiliary task to further enhance the local semantics of fused features, leading to improved overall performance. Extensive experiments demonstrate that HAPNet not only achieves state-of-the-art performance for RGB-T scene parsing across three public datasets but also shows great potential to generalize well for RGB-HHA scene parsing in indoor scenarios. Despite its superior performance over other existing approaches, the generalizability of HAPNet for RGB-X (where ``X'' represents, but is not limited to, depth, surface normal, and so forth) can be further improved. Additionally, considering scene parsing is a common functionality embedded in autonomous cars, mobile robots, and drones, the real-time performance of HAPNet is crucial, which is also a direction for possible future work.

\section{Acknowledgements}
This research was supported by the National Natural Science Foundation of China under Grant 62473288, the Fundamental Research Funds for the Central Universities, and the Xiaomi Young Talents Program.

\begin{table*}[!ht] 
	\fontsize{7.5}{13}\selectfont
	\centering
	\caption{\clr{Backbone architectures, model complexities (in total parameters, Params), and inference speeds (in frames per second, FPS) for all evaluated networks, with all performance benchmarks conducted on an NVIDIA RTX 3090 GPU. “-`` denotes that the corresponding results are unavailable.}}
	{
		\setlength{\tabcolsep}{7pt}
		\begin{tabular}{
				l|l|cc||l|l|cc
			}
			\toprule[1pt] 
			Method  &  Backbone & \clr{Params (M)} & \clr{FPS} & Method  &  Backbone & \clr{Params (M)} & \clr{FPS} \\
			\hline
			MFNet ~\cite{ha2017mfnet}  & Mini-Inception & \clr{0.72} & \clr{194.0} & CMX ~\cite{zhang2023cmx} & MiT-B4 & \clr{139.9} & \clr{11.72} \\
			RTFNet ~\cite{sun2019rtfnet}  & ResNet-152 & \clr{254.51} & \clr{26.18} & CMNeXt  ~\cite{zhang2023delivering} & MiT-B4 & \clr{119.6} & \clr{12.6} \\
			FuseSeg ~\cite{sun2020fuseseg} & DenseNet-161 & \clr{141.52} &\clr{23.85} & CRM-RGBTSeg ~\cite{shin2023complementary} &Swin-B  & \clr{193.0} & \clr{-} \\
			EGFNet ~\cite{zhou2022edge} & ResNet-152 & \clr{62.82} & \clr{8.58} & GMNet ~\cite{zhou2021gmnet} & ResNet-50 & \clr{149.8} & \clr{15.0} \\
			ABMDRNet ~\cite{zhang2021abmdrnet} & ResNet-50 & \clr{64.60} & \clr{43.46} & Omnivore ~\cite{girdhar2022omnivore} & Swin-B & \clr{95.7} &\clr{-} \\
			ECGFNet ~\cite{zhou2023embedded} & MobileNetV2 & \clr{26.06} & \clr{38.37} & TokenFusion ~\cite{wang2022multimodal} & MiT-B3 & \clr{45.9} & \clr{-}\\
			FEANet ~\cite{deng2021feanet} & Modified ResNet & \clr{255.21} & \clr{21.13} & DFormer-L ~\cite{yin2023dformer} & DFormer-L & \clr{39.0} & \clr{-}\\
			SFAF-MA ~\cite{he2023sfaf} & ResNet-152 & \clr{156.0} & \clr{21.8} & LLE-Seg ~\cite{guo2025low} & ConvNeXt V2 &\clr{-} &\clr{-} \\
			ABMDRNet+ ~\cite{zhao2023mitigating} & ResNet-50 & \clr{64.60} & \clr{40.14} & AsymFormer ~\cite{du2023asymformer} & Modified MiT-B0 & \clr{33.0} & \clr{-}\\
			CAINet ~\cite{lv2024context} & MobileNetV2 & \clr{12.16} & \clr{61.44} & SA-Gate ~\cite{chen2020bi} & ResNet-50 & \clr{110.85} & \clr{18.36} \\
			EAEFNet ~\cite{liang2023explicit} & ResNet-152 & \clr{200.4} & \clr{21.0} & OmniVec ~\cite{srivastava2024omnivec} & ViT varients & \clr{95.7} &\clr{-} \\
			
			\bottomrule[1pt]
		\end{tabular}
	}
	
	\label{tab.variance}
\end{table*}

\appendix
\setcounter{table}{0}
\section{Mathematical Details}
\label{supplement.math}

This section provides details on the mathematical derivations of the multi-head self-attention (MHSA) \cite{dosovitskiy2020image} and multi-scale deformable cross-attention (MSDA) \cite{zhu2020deformable}, which are integral components of our proposed HAPNet. MHSA is employed within the vision foundation model, while MSDA is utilized in the progressive heterogeneous feature aggregators.
\subsection{\textbf{Multi-head attention}}
Let $\boldsymbol{X}_{q}\in \mathbb{R}^{N_q \times C}$ be the query, and let $\boldsymbol{X}_{f}\in \mathbb{R}^{N_f \times C}$ be the key and value, where $N_q$ and $N_f$ are the respective sequence lengths. The multi-head attention (MHA), including both MHSA and multi-head cross-attention, can be formulated as follows:
\begin{equation}
\boldsymbol{Q}^{i}, \boldsymbol{K}^{i}, \boldsymbol{V}^{i} = \boldsymbol{X}_{q}\boldsymbol{U}_{q}^{i},\,  \boldsymbol{X}_{f}\boldsymbol{U}_{k}^{i}, \,\boldsymbol{X}_{f}\boldsymbol{U}_{v}^{i}, \,\,\,i \in [1,n]\cap\mathbb{Z},
\label{eq:qkv_proj}
\end{equation}
\begin{equation}
\boldsymbol{A}^{i} = \op{Softmax}(\frac{\boldsymbol{Q}^{i}(\boldsymbol{K}^{i})^\top}{\sqrt{C_{v}}})
\end{equation}
\begin{equation}
\text{MHA}(\boldsymbol{X}_{q}, \boldsymbol{X}_{f})= \op{Concat}([\boldsymbol{A}^{1}\boldsymbol{V}^{1};\dots;\boldsymbol{A}^{n}\boldsymbol{V}^{n}])\boldsymbol{U}_{m},
\label{eq:multihead_attn}
\end{equation}
where $\boldsymbol{U}_{q}^{i}\in \mathbb{R}^{C \times C_v}$, $ \boldsymbol{U}_{k}^{i}\in \mathbb{R}^{C \times C_v}$ and $\boldsymbol{U}_{v}^{i} \in \mathbb{R}^{C \times C_v}$ denotes the linear projections of the query, key, and value in the $i$-th attention head, respectively, $\boldsymbol{A}^{i}$ stores attention weights, and $\boldsymbol{U}_{m} \in \mathbb{R}^{n\cdot C_v \times C} $ represents the linear projection of $n$ attention operations ($C_v = \frac{C}{n}$ by default). $\op{Concat}([\cdot; \dots; \cdot])$ denotes the feature concatenation operation performed on the results from $n$ attention operations.

\subsection{\textbf{Deformable attention}} 
Compared to standard MHA, the deformable attention (DA) only attends a small set of points from the feature map as the reference points instead of the whole feature map, which reduces the computation complexity while maintaining competitive performance. The sampling process of DA can be achieved by predicting the sampling offsets and attention weights $\boldsymbol{A}^{i}$ through linear projections of the query $\boldsymbol{X}_q$. Let $\boldsymbol{p}_q = (h, w)$ index a 2-D reference point on the resized $\boldsymbol{X}_q$, where $\boldsymbol{X}_q$ has been restored to its original 2-D spatial dimensions. At this reference point, $\boldsymbol{X}_q(\boldsymbol{p}_q) \in \mathbb{R}^{1 \times C}$ represents the feature vector corresponding to position $\boldsymbol{p}_q$. The DA operation can be formulated as:
\begin{equation}
\op{DA}^{i}(\boldsymbol{X}_{q}(\boldsymbol{p}_q), \boldsymbol{X}_{f}) = \sum_{k=1}^{K} \boldsymbol{A}^{i}_{k,\boldsymbol{p}_q} \cdot \boldsymbol{X}_{f}(\boldsymbol{p}_{q} + \Delta^{i}_{k, \boldsymbol{p}_{q}})\boldsymbol{U}_{v}^{i},
\label{eq:single_deform_attn}
\end{equation}
where $k$ indexes the sampled keys, and $K$ is the total number of sampled keys. $\Delta^{i}_{k, \boldsymbol{p}_{q}}$ and $\boldsymbol{A}^{i}_{k,\boldsymbol{p}_{q}}$ denote the sampling offsets and attention weights of the $k$-th sampling point on $\boldsymbol{p}_q$ in the $i$-th attention head, respectively. The attention weights $\boldsymbol{A}^{i}_{k,\boldsymbol{p}_{q}}$ are normalized to satisfy $\sum_{k=1}^{K} \boldsymbol{A}^{i}_{k,\boldsymbol{p}_{q}} = 1$.

Let $\mathcal{X}_{f} = \{ \boldsymbol{X}^l_f\}_{l=1}^{L}$ be the multi-scale feature maps, where $\boldsymbol{X}^l_f \in \mathbb{R}^{H_l \times W_l \times C}$, and let $\boldsymbol{\hat{p}}_q \in [0,1]^2$ be the normalized coordinate of the reference point for the query $\boldsymbol{X}_q$. The $\op{DA}$ process can be extended to its MSDA form as follows:
\begin{equation}
\op{MSDA}^{i}(\boldsymbol{X}_{q}(\boldsymbol{\hat{p}}_q), \mathcal{X}_{f}) = \sum_{l=1}^{L} \sum_{k=1}^{K} \boldsymbol{A}^{i,l}_{k,\boldsymbol{p}_q} \cdot \boldsymbol{X}^l_{f}(\phi_{l}(\boldsymbol{\hat{p}}_{q}) + \Delta^{i,l}_{k, \boldsymbol{p}_{q}})\boldsymbol{U}_{v}^{i},
\label{eq:ms_deform_attn}
\end{equation}
where $l$ indexes the feature map level, and $k$ indexes the sampling point. $\Delta^{i,l}_{k, \boldsymbol{p}_{q}}$ and $\boldsymbol{A}^{i,l}_{k,\boldsymbol{p}_{q}}$ denote the corresponding sampling offsets and attention weights, respectively, for the $k$-th sampling point in the $l$-th feature level and the $i$-th attention head. The attention weights $\boldsymbol{A}^{i,l}_{k,\boldsymbol{p}_{q}}$ are normalized to satisfy $\sum_{l=1}^{L} \sum_{k=1}^{K} \boldsymbol{A}^{i,l}_{k,\boldsymbol{p}_q} = 1$. The transformation $\phi_{l}(\cdot)$ rescales the $\boldsymbol{\hat{p}}_q$ in (\ref{eq:ms_deform_attn}) to the original spatial size of the $l$-th feature level.

\section{Supplementary Content for Tables}
This section provides the detailed \clr{backbone configurations, model complexities, and inference speeds for all methods compared with HAPNet in this study, as shown in Table \ref{tab.variance}.}

\normalem

\bibliographystyle{elsarticle-num} 
\bibliography{ref.bib}

\end{document}